\pdfoutput=1

\documentclass[11pt]{article}

\usepackage{acl}

\usepackage{times}
\usepackage{latexsym}

\usepackage[T1]{fontenc}

\usepackage[utf8]{inputenc}

\usepackage{microtype}
\usepackage{latexsym}

\usepackage{booktabs}
\usepackage{multirow}

\usepackage{graphicx} 
\usepackage{caption,subcaption}

\usepackage{comment}
\usepackage{todonotes}
\usepackage{tikz}
\usepackage{amsmath}
\usepackage{breqn}
\usepackage{amssymb}
\usepackage{xcolor}
\usepackage[linesnumbered,ruled,vlined]{algorithm2e}

\SetCommentSty{mycommfont}

\SetKwInput{KwInput}{Input}                
\SetKwInput{KwOutput}{Output}              

\usepackage{pifont}
\title{Multilingual Extraction and Categorization of Lexical Collocations with Graph-aware Transformers}

 \author{Luis Espinosa-Anke$^\dagger$ Alexander Shvets${^\heartsuit}$  Alireza Mohammadshahi${^\diamondsuit} ^{\spadesuit}$ \\ {\bf James Henderson}${^\diamondsuit}$ {\bf Leo Wanner}${^\clubsuit} ^{\heartsuit}$\\
 ${^\dagger}$CardiffNLP (Cardiff University) - AMPLYFI $^\heartsuit$TALN Group, Universitat Pompeu Fabra \\
 $^\diamondsuit$Idiap Research Institute $^\spadesuit$EPFL
 $^\clubsuit$ICREA\\
 \texttt{espinosa-ankel@cardiff.ac.uk}\\\texttt{\{alexander.shvets,leo.wanner\}@upf.edu}\\\texttt{\{alireza.mohammadshahi,james.henderson\}@idiap.ch}
}

\begin{document}
\maketitle
\begin{abstract}

Recognizing and categorizing lexical collocations in context is useful for language learning, dictionary compilation and downstream NLP. However, it is a challenging task due to the varying degrees of frozenness lexical collocations exhibit. 
In this paper, we put forward a sequence tagging BERT-based model enhanced with a graph-aware transformer architecture, which we evaluate on the task of collocation recognition in context. Our results suggest that explicitly encoding syntactic dependencies in the model architecture is helpful, and provide insights on differences in collocation typification in English, Spanish and French.\footnote{Data and code are available at \\\url{https://github.com/TalnUPF/graph-aware-collocation-recognition}.} 

\end{abstract}

\section{Introduction}
\label{sec:intro}


Native speech is idiosyncratic. 
Of special prominence are syntactically-bound restricted binary co-occurrences of lexical items, in which one of the items conditions the selection of the other item. Consider a CNN sports headline from 02/15/2021:

\begin{quote}
Rafael Nadal eases into Australian Open quarterfinals, remains on course for record-breaking grand slam (cnn.com).
\end{quote}



\noindent In  this short headline, we see already three of such co-oc\-cur\-rences: {\it ease} [{\it into}] {\it quarterfinals}, {\it remain} [{\it on}] {\it course}, and {\it record-breaking grand slam}. {\it Quarterfinals} conditions the selection of [{\it to}] {\it ease} [{\it into}], {\it course} of {\it remain} [{\it on}], and {\it grand slam} of {\it record-breaking}. The idiosyncrasy of these co-occurrences becomes obvious when we look at them from a multilingual angle. Thus, in French, instead of the literal translation of  {\it ease} [{\it into}], we would use {\it se qualifier} `qualify [oneself]', in Spanish, {\it remain} [{\it on}] will be translated as {\it seguir} [{\it en}] `continue in', and in Italian {\it record-breaking} will be {\it da record}, lit. `of record' -- while the translation of {\it quarterfinals}, {\it course}, and {\it grand slam} will be literal. In lexicology, such binary co-occurrences are referred to as {\it collocations} \cite{hausmann1985kollokationen,Cowie94,Melcuk95-phras,Kilgarriff06}, with the conditioning item called the {\it base} and the conditioned item the {\it collocate}. Collocations in this sense are of high relevance to second language learning, lexicography and NLP alike, and constitute a challenge for computational models because of their heterogeneity in terms of idiosyncrasy and degree of semantic composition \cite{Melcuk95-phras}.  

Research in NLP has already addressed a number of collocation-related tasks, in particular: {\bf (1)} 
 collocation error detection, categorization, and correction in writings of second language learners \cite{ferraro2011collocations,wanner2013annotation,ferraro2014towards,rodriguez2015classification};  {\bf (2)} 
 creation of collocation-enriched lexical resources  \cite{espinosa2016extending,maru2019syntagnet,di2019verbatlas}; {\bf (3)} 
 use of knowledge on collocations in downstream NLP tasks, among them, e.g., machine translation \cite{Seretan14}, word sense disambiguation \cite{maru2019syntagnet}, natural language generation \cite{WannerBateman90-nlg}, or semantic role labeling \cite{scozzafava2020personalized}; {\bf (4)} 
 probes involving collocations for understanding to which extent language models are able to identify non-compositional meanings \cite{shwartz2019still,garcia2021probing}; and {\bf (5)} detection and categorization of collocations with respect to their semantics \cite{wanner2006making,espinosa-anke-etal-2019-collocation,Levine-etal20,anke2021evaluating}. It is this last task which is the focus of this paper.


In general, collocation identification and categorization tend to be treated as two disjoint tasks. Most of the research deals only with collocation identification \cite{smadja1993retrieving,Lin99,pecina2006combining,bouma2009normalized,Dinu-etal14,Levine-etal20}. Some  works deal with the categorization of manually precompiled lists of collocations, either in isolation \cite{Wanner-nle2004,wanner2006making,espinosa-anke-etal-2019-collocation} or with their original sentence-level contextual information \cite{Wanner-etal17-ijl}. 
Only a few works in the early phase of the neural network era of NLP address the problem of collocation identification and semantic categorization as a joint task in monolingual settings \cite{rodriguez2015classification,espinosa2016extending}. Accordingly, the performance of the models put forward in these works is still rather low. In this paper, we propose a sequence tagging framework for simultaneous collocation identification and categorization, with respect to the 
taxonomy of {\it lexical functions} (LFs) \cite{Melcuk96-lfbook}. The proposed framework is based on mono- and multilingual BERT-based sequence taggers, which are enhanced by a Graph-aware Transformer~\cite{mohammadshahi-henderson-2020-graph,10.1162/tacl_a_00358} in order to ensure that the specific syntactic dependencies between the base and the collocate are taken into account. 
The sequence taggers are executed as part of a multitask learning setup, which is complemented by a sentence classification task, which predicts the occurrence of an instance of a specific LF instance in the sentence under consideration. Our results for English, French and Spanish show the flexibility of our framework and shed light on the multilingual idiosyncrasies of  collocations.

\section{Background on Collocations}
\label{sec:backgr}






Although widely used in lexicology in the sense defined above, the term {\it collocation} is ambiguous in linguistics. As introduced by \citet{Firth57}, it refers to common word co-occurrences in discourse in general. Thus, {\it cast} and {\it vote}, {\it strong} and {\it tea}, but also {\it public} and {\it sector}, {\it night} and {\it porter}, {\it supermarket} and {\it price} form collocations in English in Firth's sense. In computational linguistics, Firth's definition has been taken up, e.g., by \cite{ChurchHanks89,Lin99,Evert07,Pecina08,bouma2009normalized,Dinu-etal14,Levine-etal20}. To avoid 
confusion between the two different senses, \citet{Krenn00-thesis} proposed to use
the narrower term {\it lexical collocation} to refer to restricted binary lexical item co-occurrences. In what follows, we will use this term to refer to the definition underlying our work.

%

Lexical collocations can be typified with respect to the meaning of the collocate and the syntactic structure formed by the base and the collocate. Practical collocations dictionaries such as, e.g., the {\it Oxford Collocations Dictionary}\footnote{ https://www.freecollocation.com/} or the {\it McMillan Collocations Dictionary}\footnote{https://www.macmillandictionary.com/collocations}, already offer a coarse-grained semantic typification. However, their typification still does not make a distinction between, e.g., {\it control} and {\it cut} in co-occurrence with {\it expenditure} or between {\it cavernous} and {\it palatial} in co-occurrence with {\it room} --- distinctions which are essential in the context of both second language learning and NLP. To the best of our knowledge, {\it lexical Functions} (LFs) \cite{Melcuk96-lfbook} are the most fine-grained taxonomy of lexical collocations.

\begin{table}[!t]
\resizebox{\columnwidth}{!}{
\centering
\begin{tabular}{p{0.25\textwidth}p{0.26\textwidth}p{0.14\textwidth}}\hline
relation & example & LF label\\ \hline
intense & {\it heavy}$_{\rm C}$ $\sim$ {\it smoker}$_{\rm B}$   & Magn\\
minor&  {\it occasional}$_{\rm C}$ $\sim$ {\it smoker}$_{\rm B}$   &  AntiMagn \\
genuine & {\it legitimate}$_{\rm C}$ $\sim$ {\it demand}$_{\rm B}$   & Ver\\
non-genuine &  {\it illegitimate$_{\rm C}$} $\sim$ {\it demand$_{\rm B}$} &  AntiVer \\ \hline
Increase.existence & {\it temperature}$_{\rm B}$ $\sim$ {\it rise}$_{\rm C}$ & IncepPredPlus \\
End.existence & {\it fire}$_{\rm B}$ $\sim$ {\it go out}$_{\rm C}$ & FinFunc0\\
A0.Come.to.effect & {\it avalanche}$_{\rm B}$ $\sim$ {\it strike}$_{\rm C}$ & Fact0\\
\hline
A0/A1.Cause.existence & {\it raise}$_{\rm C}$ $\sim$ {\it hope}$_{\rm B}$ & CausFunc0\\
A0/A1.Cause.function &  {\it start}$_{\rm C}$ $\sim$ {\it engine}$_{\rm B}$  & CausFact0 \\
Cause.decrease & {\it relieve}$_{\rm C}$ $\sim$ {\it tension}$_{\rm B}$ & CausPredMinus\\ 
A0/A1.Cause.involvement & {\it raise}$_{\rm C}$ {\it hope}$_{\rm B}$ [{\it in}]&  CausFunc1\\ 
Emit.sound & {\it elephant}$_{\rm B}$ $\sim$ {\it trumpet}$_{\rm C}$ & Son \\
\hline
A0/A1.act & {\it lend}$_{\rm C}$ $\sim$ {\it support}$_{\rm B}$   & Oper1\\
A0/A1.begin.act &  {\it gain}$_{\rm C}$ $\sim$ {\it impression}$_{\rm B}$  & IncepOper1 \\
A0.end.act & {\it withdraw}$_{\rm C}$ $\sim$ {\it support}$_{\rm B}$  & FinOper1\\
A0/A1.Act.acc.expectation   & {\it prove}$_{\rm C}$ $\sim$ {\it accusation}$_{\rm B}$ & Real1 \\
\hline
A2.Act.acc.expectation  & {\it enjoy}$_{\rm C}$ $\sim$  {\it support}$_{\rm B}$  & Real2\\
A2.Act.x.expectation & {\it betray}$_{\rm C}$ $\sim$  {\it trust}$_{\rm B}$ & AntiReal2 \\ 
\hline
    \end{tabular}
    }
    \caption{LF relations used in this paper. `A$_{\rm i}$' refer to AMR argument labels \cite{Banarescu-etal13}.}
    \label{tab:lfs}
\end{table}

A lexical function (LF) is defined as a function $f(B)$ that delivers for a base $B$ a set of synonymous collocates that express the meaning of $f$.   LFs are assigned Latin abbreviations as labels; cf., e.g., ``Oper1'' (``operare'' `perform'): Oper1({\it condolences}) = \{{\it convey}, {\it express}, {\it extend}\}; ``Magn'' (``magnum'' `big'/`intense'): Magn({\it grief}) = \{{\it deep}, {\it inconsolable}, {\it great}, \ldots\}.
Each LF can also be considered as a specific lexico-semantic relation between the base and the collocate of a collocation in question \cite{Evens88}. Table \ref{tab:lfs} displays the  subset of the relations we experiment with, along with their corresponding LF names and illustrative examples.

As seen in Table \ref{tab:lfs}, 
where pertinent, an LF label also encodes the subcategorization structure of the base+collocate combination; cf., e.g., FinFunc0, Oper1, Real2, etc. Thus, the index `1' in Oper1 encodes the information that the first argument of the base (A0/A1) is realized as grammatical subject and the base itself as object; the `2' in Real2 encodes the realization of the second argument of the base (A2) as grammatical subject and the base as object; etc. This generic structure translates into a number of {\it Universal Dependency} (UD) patterns.

\section{Related Work}
\label{sec:relatedwork}






Previous works that consider collocations in a Firthian sense look  at word adjacency in terms of $n$-grams \cite{smadja1993retrieving}, although most often, statistical measures of co-occurrence
are used; cf. \citet{pearce2002comparative,pecina2006combining,pecina2010lexical,garcia2019comparison}. Some complement statistical measures by morphological \cite{krenn2001can,evert2001methods} and/or syntactic \cite{heid1989collocations,Lin99,seretan2006accurate} patterns. 
In view of the \textit{asymmetrical} nature of the relation between the base and the collocate, e.g., \citet{gries201350} proposes
to investigate ``directional measures'' as an addition to association measures; \citet{carlini2014improving} explicitly encode this asymmetry in terms of NPMI \cite{bouma2009normalized}, which is a normalized version of PMI; see also \cite{garcia2019comparison}. In the collocation classification task, substantial research focused on the identification of {\it Light Verb Constructions}, which are captured by the Oper- (and partially by the Real-) families of LFs; cf., e.g., \cite{Dras95,Vincze-etal2013_SVCs,Kettnerova-etal-2013,Chen-etal2016,cordeiro-candito-2019-syntax,shwartz2019still}, whereas \citet{Chung-Chi-etal09} and \citet{Wanner-etal17-ijl} focus on broad semantic collocation categories. Several works also use LFs as a collocation taxonomy. Thus, \citet{wanner2006making} leverage a  vector-based similarity metric on a subset of LFs, whereas \citet{gelbukh2012semantic} explore a suite of classical supervised ML algorithms.

More recently, word embeddings have been successfully applied in unsupervised setups, e.g.,  \citet{rodriguez2016example} use simple vector arithmetic. In supervised setups, we find, first, the ``collocate retrieval'' approach proposed by \citet{rodriguez2016semantics}, who train a linear transformation to go from a ``base'' to a ``collocate'' vector space, exploiting regularities in multilingual word embeddings \cite{mikolov2013exploiting}, and second, 
\citet{espinosa-anke-etal-2019-collocation}, who train an SVM on a dedicated relation vector space for base and collocate. Embeddings have also been used  in multilingual English/Spanish \cite{rodriguez2016semantics} and English/Portuguese/Spanish \cite{garcia2017using} LF classification. While successful, none of these approaches explicitly leveraged in the language model the crucial syntactic dependency information between base and collocate, or considered how sentence-level information could benefit the extraction task -- as we do.

\section{Graph-to-Collocation Transformer}
\label{sec:framework}


\begin{figure}[!t]
\resizebox{\columnwidth}{!}{
\includegraphics[]{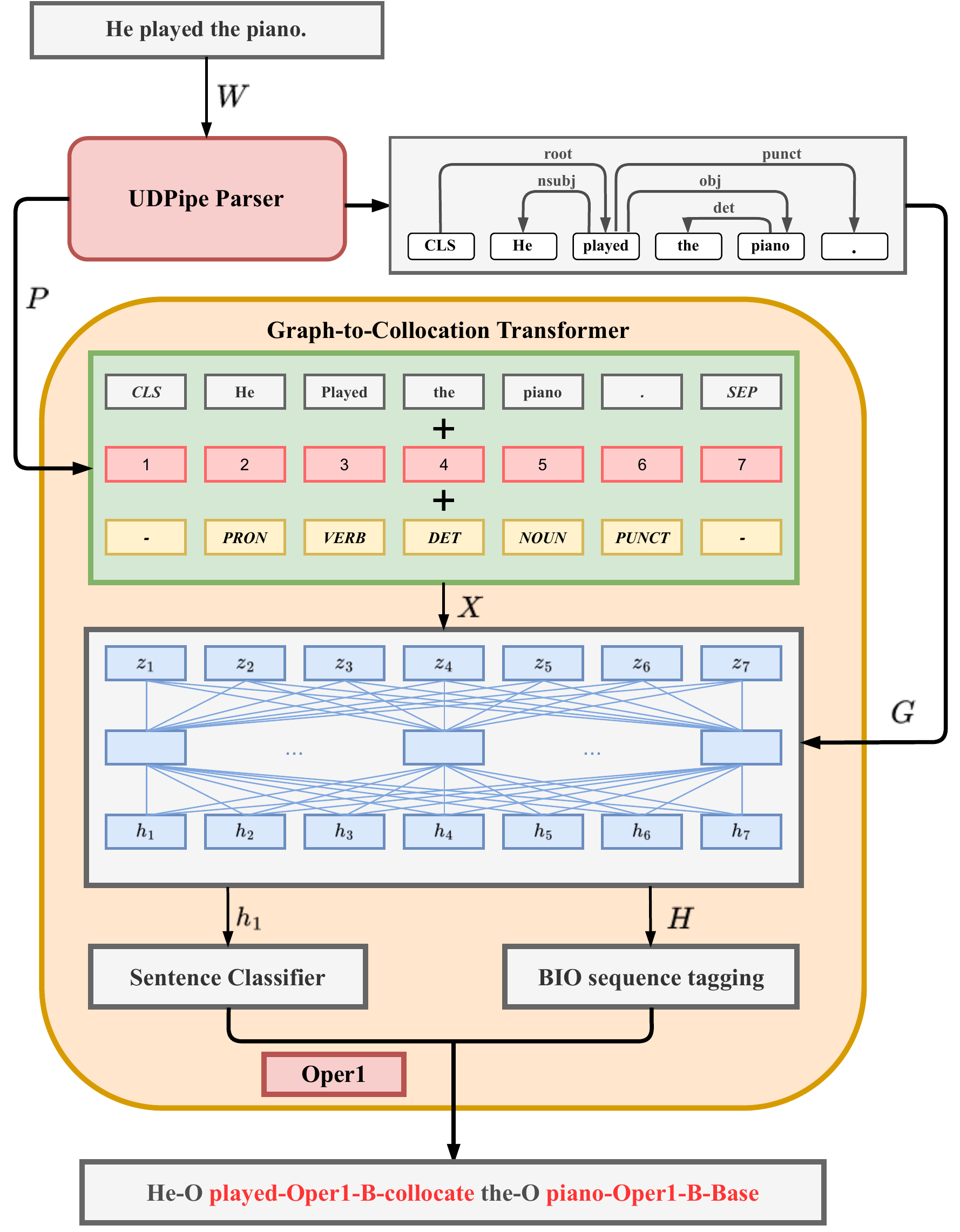}
}\caption{Graph-to-Collocation Transformer, which generates a BIO-tagged sequence given a sentence with, optionally, its parsed tree.}
\label{fig:main_model}
\end{figure}

We propose a Graph-to-Collocation Transformer (G2C-Tr) to: (1) cast collocation identification and classification as a \textbf{sequence tagging} problem: as pointed out above, lexical collocations are lexico-semantic relations, and relation extraction has been recently successfully addressed as sequence tagging \cite{Ji-etal21}; (2) \textbf{boost performance} by enabling multitask learning via joint sentence classification and LF-instance BIO tagging; and (3) capture the asymmetric \textbf{semantic and syntactic dependency} between the base and the collocate by the use of a modified attention mechanism.

The G2C-Tr is implemented as a suite of BERT-based models for joint sentence classification and sequence tagging.  The syntactic dependency graph of the sentence is input to a G2C-Tr model through its attention mechanism.
Figure~\ref{fig:main_model} illustrates the framework of our model. Given the input sentence $W=(\mathbf{w}_1,\mathbf{w}_2,...,\mathbf{w}_N)$, we first use a pre-trained dependency parser $\operatorname{DP}()$ to build the dependency graph $G$, and Part-of-Speech (PoS) tags $P=(\mathbf{p}_1,\mathbf{p}_2,...,\mathbf{p}_N)$. Due to the fact that each LF is characterized by the PoS of its lexical items and the syntactic dependency between them, this information is of significant importance. Then, G2C-Tr predicts the tagged sequence $Y=(\mathbf{y}_1,\mathbf{y}_2,...,\mathbf{y}_N)$ as follows:
\begin{equation}
  \begin{cases}
  P,G = \operatorname{DP}(W) \\
  H = \operatorname{Enc}(W,P,G) \\
  Y = \operatorname{Dec}(H)
  \end{cases}
\label{eq:main_model}
\end{equation}

\noindent where $\operatorname{Enc}()$, $\operatorname{Dec}()$ are the encoder and decoder parts of our model, described below. $H=[\mathbf{h}_1, \dots, \mathbf{h}_T]$ is the contextualised vector representation, and $T$ is the length of the tokenized sequence. The parameters of $\operatorname{DP}()$ are frozen for training. 

\subsection{Encoder}
\label{model:enc}

To compute the contextualised vector embeddings $H$, we use a modified version of the Graph-to-Graph Transformer model proposed by \citet{10.1162/tacl_a_00358} to encode both PoS tags~($P$) and the dependency graph~($G$). Let us first introduce the encoding mechanism.

\subsubsection{Input Embeddings}

Given an input sentence~($W$) with its associated PoS tags~($P$), the G2C-Tr model first computes the input embeddings~($X=(\mathbf{x}_1,\mathbf{x}_2,...,\mathbf{x}_T)$). To make it compatible with BERT~\cite{devlin-etal-2019-bert}, we append two special tokens, {\tt CLS}, and {\tt SEP} to the start and end of the tokenized sequence, respectively. The input embeddings are calculated as the summation of pre-trained token embeddings of BERT, position embeddings, and PoS tag embeddings (as shown in the green part of Figure~\ref{fig:main_model}). 

\subsubsection{Self-attention Mechanism}

Given the input embeddings~($X$), and a dependency graph~($G$), we compute the contextualised vector representations~($H$) using a modified version of the Transformer architecture. The original Transformer model~\cite{NIPS2017_3f5ee243} is composed of several Transformer layers. Each Transformer layer includes a self-attention module and a position-wise feed-forward network. Previous work~\cite{ying2021transformers,mohammadshahi-henderson-2020-graph,10.1162/tacl_a_00358,mohammadshahi2021syntaxaware} modified the attention mechanism by adding scalar biases to the attention scores~\cite{ying2021transformers}, or multiplying the query representation with relation vectors~\cite{10.1162/tacl_a_00358,mohammadshahi-henderson-2020-graph} to encode graph structures. 

Since in collocations, base and collocate are syntactically related and LFs are characterized by specific dependency relations, we modify the attention mechanism of the base transformer model to inject syntactic information. In each Transformer layer, given $Z_n=(\mathbf{z}_1,\mathbf{z}_2,...,\mathbf{z}_T)$ as the output representations of the previous layer, the attention weights are calculated as a Softmax over the attention scores $\alpha_{ij}$, defined as: 
\vspace*{-0.2cm}
\begin{dmath}
\alpha_{ij} = \frac{1}{\sqrt{3d}} \Big[ \mathbf{z}_i\boldsymbol{W^Q}(\mathbf{z}_j\boldsymbol{W^K})^T+\mathbf{z}_i\boldsymbol{W^Q}(\mathbf{r}_{ij}\boldsymbol{W^R_A})^T+\mathbf{r}_{ij}\boldsymbol{W^R_A}(\mathbf{z}_j\boldsymbol{W^K})^T \Big]
\label{eq:g2g-attn}
\end{dmath}
\vspace*{-0.2cm}

\noindent where $\boldsymbol{W^Q}, \boldsymbol{W^K} \in \mathbb{R}^{d_h \times d}$ are learned query and key parameters. $\boldsymbol{W^R_A} \in \mathbb{R}^{2|G|+1 \times d}$ is the graph relation embedding matrix, learned during training, $d_h$ is the dimension of hidden vectors, $d$ is the head dimension of self-attention module, and $|G|$ is the overal number of dependency labels. $\mathbf{r}_{ij}$ is the one-hot vector representing both the relation and direction of syntactic relation between token $\mathbf{x}_i$ and $\mathbf{x}_j$, so $\mathbf{r}_{ij}\boldsymbol{W^R_A}$ selects the embedding vector for the appropriate syntactic relation. Algorithm~\ref{alg_g2g} shows the procedure of building relation matrix $R$. Finally, we also add the graph information to the value computation of the Transformer as:

\begin{algorithm}[!t]
\DontPrintSemicolon
  
  \KwInput{Graph $G = \{(i,j,l)\},j=1,..,T$ }
  \tcc{$i$,$j$,$l$ are parent node id, dependent id and label}
  \tcc{{\tt CLS} is the root node}
  \KwOutput{Relation Matrix $R$}
  R = $\operatorname{zeros}(T,T)$ \\
  \For{$(i,j,l) \in G$}     
  { 
    $r_{i,j} = k_{l}$ \\
    $r_{j,i} = k_{l} + |G|$
  }
  \tcc{$k_l$ is the index of label $l$}

\caption{Build Relation Matrix $R$}
\label{alg_g2g}
\end{algorithm}

\vspace*{-0.2cm}
\begin{align}
\begin{split}
\mathbf{v}_i = \sum_j \frac{\exp(\alpha_{ij})}{\sum_j \exp(\alpha_{ij})}(\mathbf{z}_j \boldsymbol{W^V} + \mathbf{r}_{ij} \boldsymbol{W^R_V})
\end{split}
\label{eq:g2g-attn2}
\end{align}
\vspace*{-0.25cm}

\noindent where $\frac{\exp(\alpha_{ij})}{\sum_j \exp(\alpha_{ij})}$ is the Softmax for the attention weights, $\boldsymbol{W^V} \in \mathbb{R}^{d_h \times d}$ is the learned value matrix, $\boldsymbol{W^R_V} \in \mathbb{R}^{2|G|+1 \times d}$ is the graph embedding parameter, and $v_i$ is the output representation of the self-attention mechanism for the token $i$. To find the output representations~($H$), we use the same mechanism for position-wise feed-forward layer, and layer normalisation as proposed in  \citet{NIPS2017_3f5ee243}. \\
Intuitively, additional terms in Equation~\ref{eq:g2g-attn}~(second and third multiplications), and Equation~\ref{eq:g2g-attn2}~(second addition) add a soft bias toward the syntactic information. The model can still decide to use the injected syntactic information, or just rely on the context information~(first terms in both Equation~\ref{eq:g2g-attn} and \ref{eq:g2g-attn2}). 

\subsection{Decoder}
\label{model:dec}

BERT-based joint sentence classification and sequence tagging has already been used, e.g., for natural language understanding in the context of question answering and goal-oriented dialogue systems, where it serves for \textit{speaker intent} identification and \textit{semantic frame slot filling} \cite{chen2019bert,castellucci2019multi}. In the context of sentence classification, we can specify such a model as: 

\vspace*{-0.2cm}
\begin{align}
\begin{split}
y^i = \text{softmax} \left ( \mathbf{W}^i\mathbf{h}_1 + \mathbf{b}^i \right ),
\end{split}
\label{eq:class}
\end{align}
\vspace*{-0.2cm}

\noindent with $i$ as the index of the sentence that is to be classified, and $\mathbf{h}_1$ as the hidden state of the first pooled special token ({\tt CLS} in the case of BERT). For sequence tagging, this equation is extended such that the sequence $[\mathbf{h}_2, \dots, \mathbf{h}_T]$ is fed to word-level softmax layers:

\vspace*{-0.2cm}
\begin{align}
\begin{split}
y^s_n = \text{softmax} \left ( \mathbf{W}^i\mathbf{h}_n + \mathbf{b}_n \right ), n \in 1 \dots |W| 
\end{split}
\label{eq:seq}
\end{align}
\vspace*{-0.2cm}


\noindent where $\mathbf{h}_n$ is the hidden state corresponding to $w_n$. Finally, the joint model combines both architectures and is trained, end-to-end, by minimizing the cross-entropy loss for both tasks.


\vspace*{-0.2cm}
\begin{equation}
\begin{split}
p \left ( y^i, y^s | W \right ) = p \left ( y^i | H \right ) \prod_{n=1}^Np \left ( y_n^s| H \right ) 
\end{split}
\label{eq:joint}
\end{equation}
\vspace*{-0.20cm}

\section{Experimental setup}
\label{sec:experiments}




\subsection{Dataset Construction}
\label{sec:dataset}

We carry out experiments on English, French, and Spanish datasets constructed  from manually compiled instances of LFs.
For English and French, we start from \citet{fisas-etal-2020-collfren}.
For English, Fisas et al.'s list is enriched by 500 instances of low-resourced LFs in order to obtain a more balanced distribution of samples across different LFs; for French, we work with their original list. To obtain the LF instances for Spanish, we use the English list: for each English LF instance, we retrieve from the web via the
multilingual search index {\it ReversoContext}\footnote{https://context.reverso.net/}  its translation equivalents, which are then examined and filtered manually.

In all three lists, the bases and collocates are annotated with  PoS and lemmas. As corpora, we use the 2019 Wikipedia dumps. First, we preprocess (removing metadata and markups) and parse the dumps with the  UDPipe2.5 parsers.\footnote{\url{https://ufal.mff.cuni.cz/udpipe}} Then, we extract from the parsed dumps sentences that contain LF instances from any of our collocation lists, observing the PoS of the base and collocate and the dependency relation between them. 
 To further filter the remaining erroneous samples in which the base and the collocate items do not form a collocation, an additional manual check is performed.


The validated sentences and the collocations they contain are labeled. As sentence label, the sentence's most frequent LF or the first one in case of a draw is chosen.
In practice, this most often means that the label of the only LF instance in the sentence is chosen. For instance, in the case of CausFunc0, in the French dataset, only in 1.63\% of the cases its instances appear together with instances of other LFs in a sentence, in the Spanish dataset these are 1.85\% and in the English dataset 3.42\%. However, it should be noted that this varies from LF to LF and for some of the LFs our labeling strategy might be an oversimplification. The highest percentage of ``cohabitation'' with instances of other LFs can be observed for Oper1: in the French dataset in 7.19\% of the cases, in the Spanish dataset in 14.32\% and in the English dataset in 25.61\%. A more detailed study is necessary to identify potential correlations between different LFs.\footnote{We would like to thank an anonymous reviewer for pointing out the relevance of the correlation between LFs.}

To annotate  collocations, we use the BI labels of the BIO sequence annotation schema (`B-$<$LF$>_b$' and `I-$<$LF$>_b$' for the base, `B-$<$LF$>_c$', `I-$<$LF$>_c$' for the collocate, and `O' for other tokens) (Figure \ref{fig:main_model}). The BIO annotation facilitates a convenient labeling of multi-word elements, and the separate annotation of the base and collocate allows for flawless annotation of cases where they are not adjacent.



\begin{figure}[!h]
\resizebox{\columnwidth}{!}{
\includegraphics[]{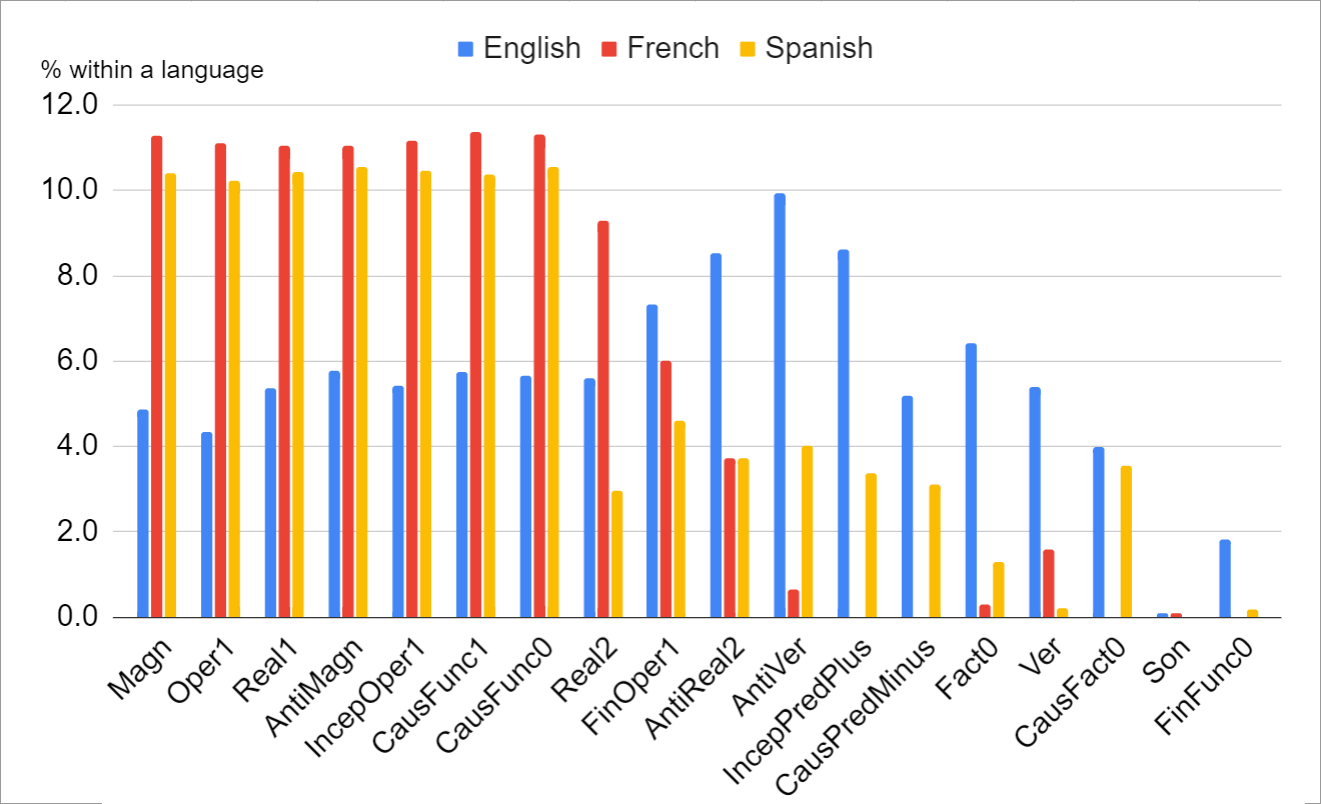}
}\caption{Distribution of examples across lexical functions within a language.}
\label{fig:lf_distribution}
\end{figure}

For the experiments, the annotated datasets are split into training, development, and test subsets in proportion 80--10--10 in terms of LF-wise unique instances, such that all occurrences of a specific instance, i.e., a specific lexical collocation, appear only in one of the subsets. Sentences with several collocations that belong to different splits are dropped. 
The distribution of samples per LF and language is shown in Figure \ref{fig:lf_distribution}. 

\subsection{Experiments}
\label{sec:exps}

In our experiments, we compare the following architectures:\footnote{In all cases, we report only results for the joint architecture, as initial experiments showed a consistent improvement with respect to a sequence tagging-only setup.} 

\begin{itemize}

\item Baseline BERT (or similar)-based models (denoted as -- in the results tables), specifically BERT-base and large \cite{devlin-etal-2019-bert}, RoBERTa-base and large \cite{liu-etal-2019-robust}; CamemBERT \cite{martin2019camembert} and RoBERTa-BNE \cite{gutierrez2021spanish} as monolingual French and Spanish models; and XLM-R for cross-lingual experiments \cite{conneau2019unsupervised}.

\item Enhanced architectures with the G2C architecture, but without access to the PoS embeddings (G2C (wo) PoS).

\item The full model, as depicted in Figure \ref{fig:main_model}, which we refer to as `G2C'.

\end{itemize}

In terms of hyperparameter tuning, we fine-tune learning rate and warmup independently for the baseline, G2C(wo)PoS and G2C English models, and fix these values for both French and Spanish. We also use early stopping on the validation set for selecting the best performing models in each configuration.



\section{Results}
\label{sec:results}


In what follows, we first present the outcome of the sentence classification and collocation extraction and categorization experiments for the three datasets and then analyze the performance with respect to the individual LFs.

\subsection{Sentence classification and collocation extraction results}
\label{sec:results:main}

Tables \ref{tab:main-res}--\ref{tab:spanish} show the performance of various joint models in their original form (marked by `--'), as well as of their G2C(wo)PoS and G2C enhanced variants. We display results on the development (`Dev*') and test sets (`Test*')  for the tasks of both sentence classification (`*SentClf') and collocation extraction (`*CollExt'). Sentence classification results are reported in terms of accuracy (there are 18 distinct LF labels), whereas 
for the collocation extraction task, we report macro F1 over correctly predicted spans. For all experiments, we report average score and standard deviation after three independent runs.

\begin{table}[h]
\Huge
\renewcommand{\arraystretch}{1.3}
\resizebox{\columnwidth}{!}{
\begin{tabular}{@{}llrrrr@{}}
\toprule
                               &                                & \multicolumn{1}{l}{DevSentClf} & \multicolumn{1}{l}{DevCollExt} & \multicolumn{1}{l}{TestSentClf} & \multicolumn{1}{l}{TestCollExt} \\ \midrule
\multirow{3}{*}{BERT$_{b}$}     & --                             & 66.86+-5.08                    & 63.21+-1.41                     &  66.04+-1.13                     & 62.95+-3.51                    \\
                               & G2C(wo)PoS                            & 61.72+-2.92 & 59.90+-1.50 & 65.18+-1.61 & 63.61+-1.25                    \\
                               & G2C                        & 64.23+-1.34 & 62.48+-0.94 & \textbf{67.25+-0.82} & 64.44+-1.12                    \\ \midrule
\multirow{3}{*}{BERT$_{l}$}    & --                             & 66.79+-1.89 & 65.69+-1.66 & 63.05+-1.23 & 61.61+-1.15                   \\
                               & G2C(wo)PoS                            & 67.58+-1.19 & 66.13+-1.48 & 66.24+-3.30 & 64.38+-3.36                  \\
                               & G2C                        & \textbf{70.30+-1.89} & \textbf{68.82+-0.86} & 64.57+-3.60 & 62.70+-3.74                    \\ \midrule
\multirow{3}{*}{RoBERTa$_{b}$}  & --                             & 58.09+-0.49                              & 55.93+-1.52                              & 60.96+-1.72                               & 59.20+-3.31                               \\
                               & G2C(wo)PoS                            & 59.89+-1.06                              & 58.05+-0.40                              & 62.51+-0.37                               & 62.17+-0.74          \\
                               & G2C                        & 59.76+-0.78                              & 58.00+-0.35                              & 62.17+-0.67                               & 61.90+-0.97                                                      \\ \midrule
\multirow{3}{*}{RoBERTa$_{l}$} & --                             & 67.47+-2.77                              & 66.97+-1.14                              &  65.55+-0.83                             & 64.79+-3.12                               \\
                               & G2C(wo)PoS                            & 67.40+-3.49                              & 67.97+-4.77                              & 65.95+-2.44                               & 64.84+-1.29          \\
                               & G2C                        & 61.71+-2.57                              & 59.85+-2.95                              & 65.10+-3.24                               & \textbf{64.98+-2.85}                               \\ \bottomrule
\end{tabular}
}
\caption{Main results for the English dataset, comparing BERT and RoBERTa, in their base ($_{b}$) and large ($_{l}$) variants, and in vanilla (--) and G2C versions.}
\label{tab:main-res}
\end{table}

The results let us conclude, firstly, that the proposed model is considerably more competitive for the task of the compilation of LF-classified collocation resources than competitive baselines. Secondly,  incorporating the G2C architecture contributes to an improvement in performance across the board, for all three languages and for most of the models. Thus, for English we see that BERT base sees an improvement of 1 and 2 points in the sentence classification and sequence labeling results on both the development and test sets, with the improvement on BERT large and RoBERTa base being even more pronounced. RoBERTa large seems to be the model that benefits least from G2C architectures in relative terms, although comparatively, this model is the best performing one on the collocation extraction task on the test set.


With respect to the experiments on French, we can observe that the French camemBERT model does not profit from an enhancement with G2C(wo)PoS; just on the contrary, for the collocation extraction task, performance drops significantly when expanded with either of the G2C variants. This is not the case for XLM-R with its different training variants; its performance is largely maintained in collocation extraction with G2C regimes. The best performance is achieved when XLM-R is enhanced with G2C and trained on both French and English. This also true for the sentence classification task. It is interesting to observe that when trained on English, XLM shows on the development set 
a higher performance than its extensions for both tasks.

\begin{table}[!h]
\Huge
\renewcommand{\arraystretch}{1.3}
\resizebox{\columnwidth}{!}{
\begin{tabular}{llrrrr}
                                                        \toprule                    &         & \multicolumn{1}{l}{DevSentClf} & \multicolumn{1}{l}{DevCollExt} & \multicolumn{1}{l}{TestSentClf} & \multicolumn{1}{l}{TestCollExt} \\ \midrule
\multirow{3}{*}{\begin{tabular}[c]{@{}l@{}}camembert\\ Tr: FR\end{tabular}} & --      & 66.69+-2.37 & 62.18+-3.32 & 54.52+-3.10 & 51.96+-2.78 \\
                                                                            & G2C(wo)PoS     & 64.38+-1.79 & 38.99+-2.45 & 50.43+-3.09 & 30.63+-3.50 \\
                                                                            & G2C & 63.60+-1.33 & 39.36+-6.38 & 50.16+-0.46 & 30.62+-5.24 \\\midrule
\multirow{3}{*}{\begin{tabular}[c]{@{}l@{}}XLM-r\\ Tr: FR\end{tabular}}     & --      & 62.22+-2.40 & 59.30+-5.04 & 56.38+-3.47 & 55.23+-3.33 \\
                                                                            & G2C(wo)PoS     & 67.08+-4.07 & 64.32+-6.20 & 58.41+-3.51 & 56.97+-2.24 \\
                                                                            & G2C & 64.63+-5.93 & 61.05+-5.57 & 56.99+-1.54 & 55.92+-1.78 \\\midrule
\multirow{3}{*}{\begin{tabular}[c]{@{}l@{}}XLM-r\\ Tr: EN\end{tabular}}  & --      & \textbf{67.18+-1.99}                              & \textbf{64.54+-5.65}                              & 54.60+-0.69                               & 52.84+-0.04                               \\
                                                                            & G2C(wo)PoS     & 65.86+-1.83                              & 64.42+-6.84                              & 54.23+-3.12                               & 50.96+-1.05                               \\
                                                                            & G2C & 65.46+-1.49                              & 64.09+-1.03                              & 55.20+-3.62                               & 52.43+-3.77       \\ \midrule\multirow{3}{*}{\begin{tabular}[c]{@{}l@{}}XLM-r\\ Tr: FR+EN\end{tabular}}  & --      & 63.07+-2.46                              & 61.59+-1.88                              & 63.35+-2.15                               & 61.32+-1.27                               \\
                                                                            & G2C(wo)PoS     & 64.40+-0.34                              & 63.88+-1.27                              & 64.95+-0.85                               & 63.55+-0.84                               \\
                                                                            & G2C & 62.02+-1.53                              & 61.03+-3.72                              & \textbf{66.48+-1.55}                               & \textbf{64.96+-2.02}       \\ \bottomrule
\end{tabular}
}
\caption{Main results for French, comparing the monolingual model CamemBERT with XLM-R variants trained on different slices of the dataset, and G2C(wo)PoS-based extensions.}
\label{tab:french}
\end{table}

For Spanish, the performance of the monolingual RoBERTa is in clear contrast to its performance on English. Although it  somewhat profits from the G2C enhancement, it seems to underperform compared to XLM-R (which is not the case for English). The reason might be the corpus on which it has been pre-trained (the National Library of Spain corpus) or under-tuning of the set of hyperparameters, which we optimized on the English dataset.
We also experiment with XLM-R, trained also only on the Spanish monolingual data (Tr: ES), as well as on the English training set (Tr: EN), and both combined (Tr: ES+EN). Surprisingly enough, XLM-R (stand-alone and G2C+POS-enhanced) performs somewhat better on the test set for both sentence classification and LF-classification when trained on English than when trained on Spanish. In general, the increase in performance provided by the multilingual setting becomes apparent\footnote{We leave for future work an analysis of whether these results can be fully attributed to multilingual transfer, to having access to more training data, or to a combination of the two.}, with the G2C model yielding the best results in 3 out of 4 metrics. The best test results of a non-G2C-enhanced model on the collocation extraction task are almost 10 points below the G2Cs models. Moreover, combining both EN and ES training sets into a multilingual language model results in an increase of 6\% F1 score. Finally, the differences in the performance of sentence classification and collocation extraction for all three datasets suggest that the predicted sentence label does not always match the label predicted by the BIO-tagger. However, since our primary intention was to use the sentence classifier as an auxiliary task that boosts the performance of the BIO-tagger in a multitask learning setup, we did not analyze the behavior of the sentence classifier and these mismatches in detail.

\begin{table}[!t]
\Huge
\renewcommand{\arraystretch}{1.3}
\resizebox{\columnwidth}{!}{
\begin{tabular}{@{}llrrrr@{}} \toprule
                                                                                      &         & DevSentClf & DevCollExt & TestSentClf & TestCollExt \\ \midrule
\multirow{3}{*}{\begin{tabular}[c]{@{}l@{}}RoBERTa$_{es}$\\ Tr: ES\end{tabular}} & --      & 34.42+-0.65 & 26.65+-1.20 & 37.90+-0.67 & 27.94+-0.16           \\
                                                                                      & G2C(wo)PoS     & 35.62+-1.90 & 28.42+-2.20 & 38.60+-1.33 & 29.73+-2.05           \\ 
                                                                                      & G2C & 37.60+-3.14 & 31.20+-1.63 & 40.49+-0.84 & 31.20+-5.47           \\\midrule
\multirow{3}{*}{\begin{tabular}[c]{@{}l@{}}XLM-r\\ Tr: ES\end{tabular}}            & --      & 66.44+-1.02          & 62.77+-0.01          & 52.99+-0.29           & 51.57+-0.12           \\
                                                                                      & G2C(wo)PoS     & 68.69+-1.96          & 66.08+-1.95          & 54.96+-0.35           & 53.74+-0.42           \\
                                                                                      & G2C & 63.96+-5.06          & 65.32+-2.20          & 56.42+-0.84           & 55.07+-0.71           
                     \\\midrule
\multirow{3}{*}{\begin{tabular}[c]{@{}l@{}}XLM-r\\ Tr: EN\end{tabular}}            & --      & 65.02+-1.61          & 63.16+-1.93          & 60.56+-0.52           & 56.95+-2.48           \\
                                                                                      & G2C(wo)PoS     & 63.00+-0.72          & 62.21+-0.67          & 58.82+-1.41           & 57.90+-0.62           \\
                                                                                      & G2C & 62.54+-0.45          & 61.37+-0.48          & 57.65+-1.81           & 54.50+-1.57                                                                      \\\midrule
\multirow{3}{*}{\begin{tabular}[c]{@{}l@{}}XLM-r\\ Tr: ES+EN\end{tabular}}         & --      & 65.91+-0.13          & 62.73+-0.59          & 64.26+-1.97           & 63.37+-0.72           \\
                                                                                      & G2C(wo)PoS     & 74.18+-1.01          & 71.20+-0.88          & 75.42+-0.02           & \textbf{72.89+-0.07}            \\
                                                                                      & G2C & \textbf{74.52+-0.18}          & \textbf{71.64+-0.01}          & \textbf{75.55+-0.18}           & 72.18+-0.92      \\    \bottomrule
\end{tabular}
}
\caption{Main results for Spanish, comparing the monolingual model RoBERTa-bne with XLM-R variants trained on different slices of the dataset, and G2C(wo)PoS-based extensions.}
\label{tab:spanish}
\end{table}

\begin{figure*}[!tb]
    \centering 
\begin{subfigure}{0.30\textwidth}
  \includegraphics[width=\linewidth]{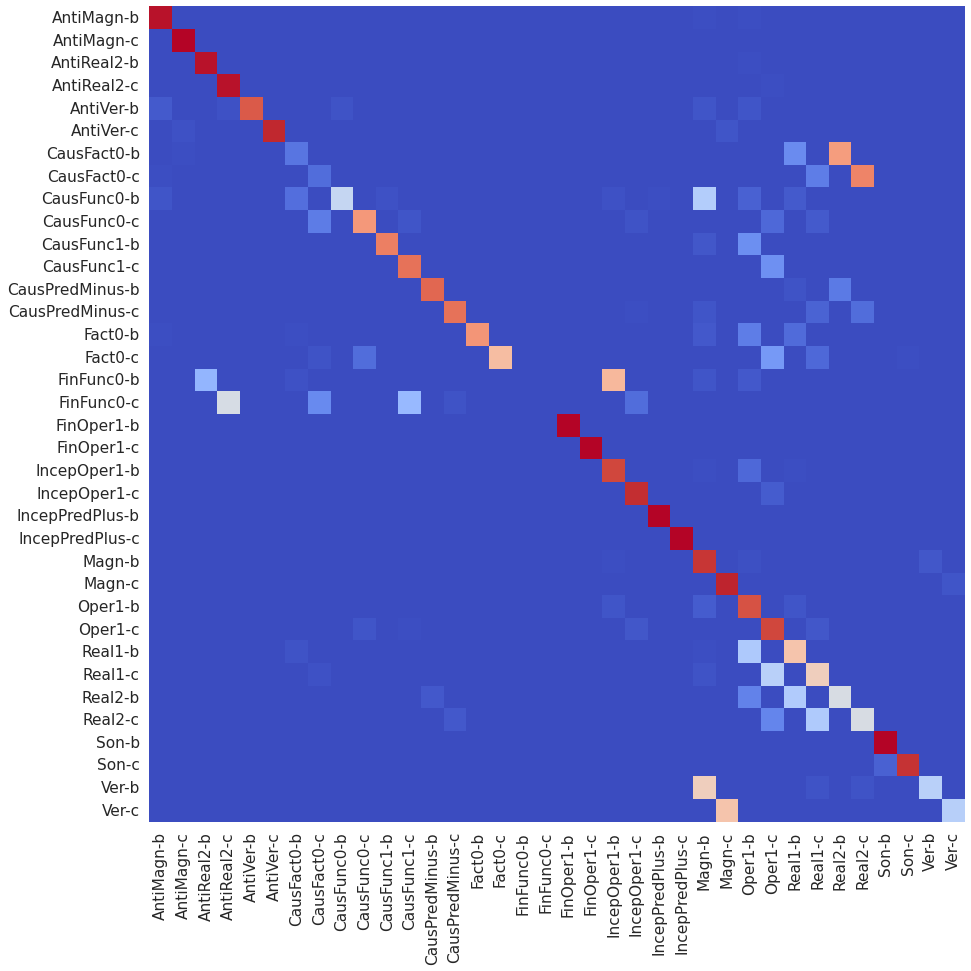}
  \caption{English.}
  \label{fig:1}
\end{subfigure}\hfil 
\begin{subfigure}{0.30\textwidth}
  \includegraphics[width=\linewidth]{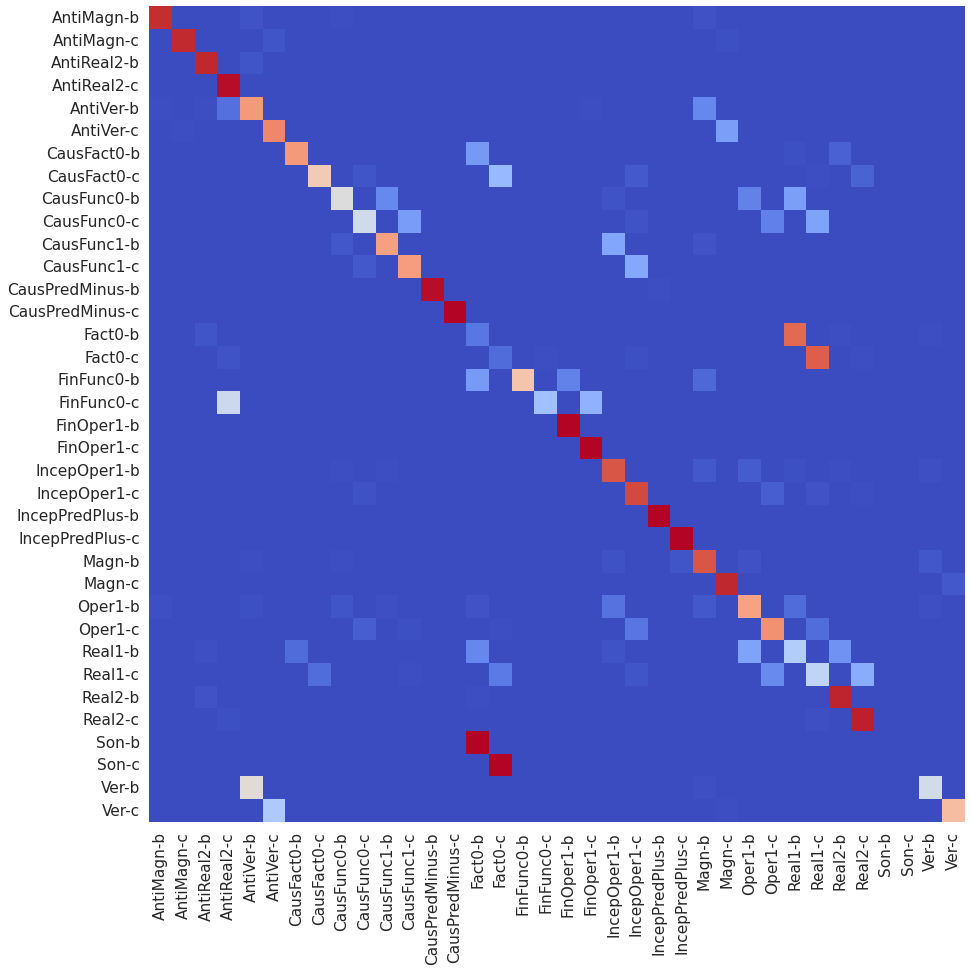}
  \caption{Spanish.}
  \label{fig:2}
\end{subfigure}\hfil 
\begin{subfigure}{0.30\textwidth}
  \includegraphics[width=\linewidth]{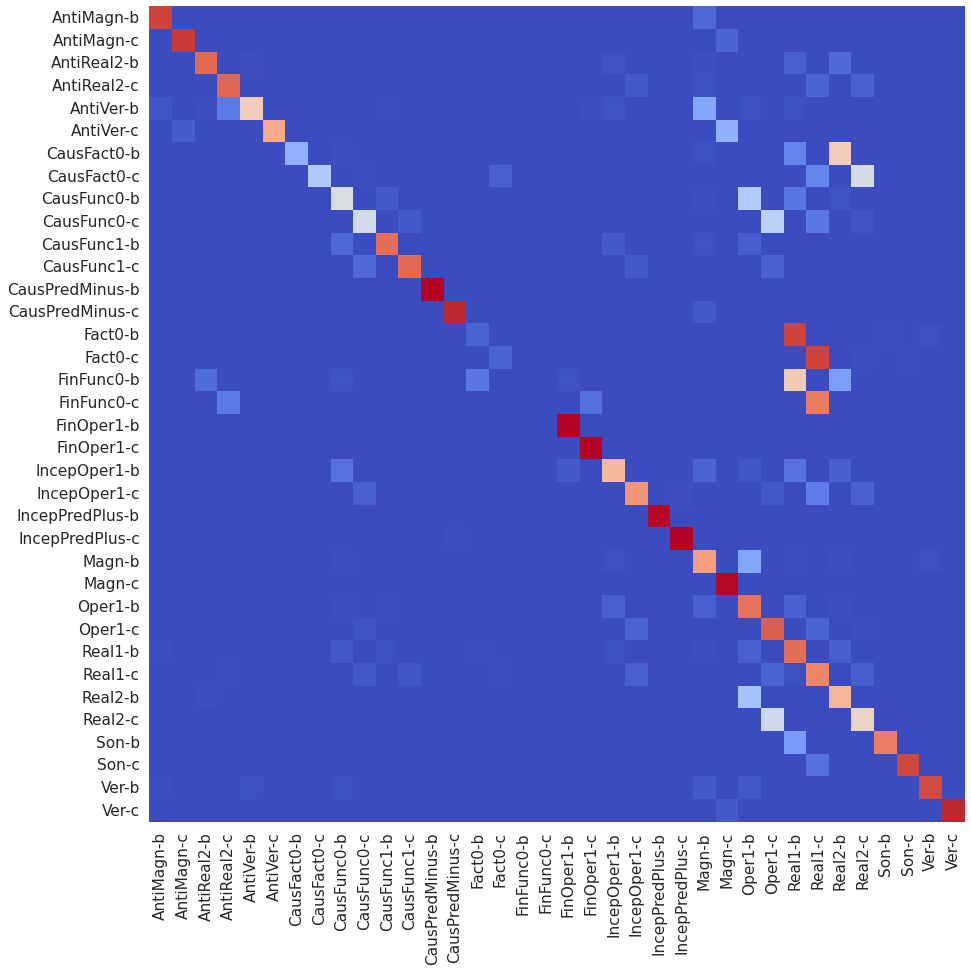}
  \caption{French.}
  \label{fig:3}
\end{subfigure}

\begin{subfigure}{0.30\textwidth}
  \includegraphics[width=\linewidth]{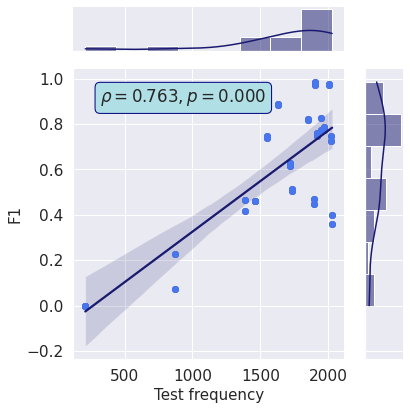}
  \caption{English.}
  \label{fig:4}
\end{subfigure}\hfil 
\begin{subfigure}{0.30\textwidth}
  \includegraphics[width=\linewidth]{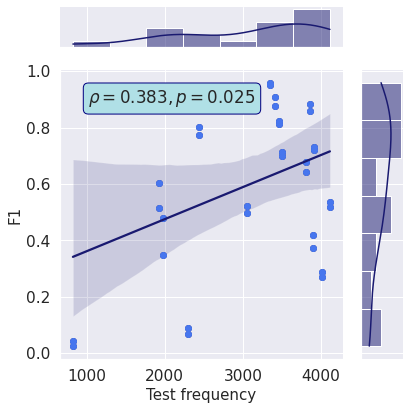}
  \caption{Spanish.}
  \label{fig:5}
\end{subfigure}\hfil 
\begin{subfigure}{0.30\textwidth}
  \includegraphics[width=\linewidth]{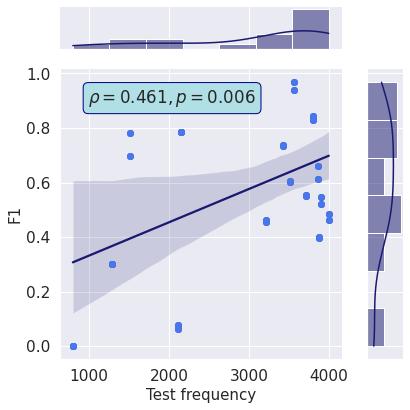}
  \caption{French.}
  \label{fig:6}
\end{subfigure}
\caption{LF analysis visualization. Top row shows confusion matrices for the three languages under study, for all LFs and their corresponding base and collocate label. Bottom row shows scatter plot where we show frequency in the x axis, and F1 score in the y axis, again, for each LF.}
\label{fig:lf-analysis}
\end{figure*}

\subsection{Lexical Function analysis}
\label{sec:results:lfs}


\begin{table}[!th]
\Huge
\renewcommand{\arraystretch}{1.3}
\resizebox{\columnwidth}{!}{
\begin{tabular}{@{}l|rrr|rrr|rrr@{}}
\toprule
                 & \multicolumn{3}{c|}{EN}                                                          & \multicolumn{3}{c|}{ES}                                                                            & \multicolumn{3}{c}{FR}                                                                            \\ \midrule
                 & \multicolumn{1}{l}{\textbf{P}} & \multicolumn{1}{l}{\textbf{R}} & \multicolumn{1}{l|}{\textbf{F1}} & \multicolumn{1}{l}{\textbf{P}} & \multicolumn{1}{l}{\textbf{R}} & \multicolumn{1}{l|}{\textbf{F1}} & \multicolumn{1}{l}{\textbf{P}} & \multicolumn{1}{l}{\textbf{R}} & \multicolumn{1}{l}{\textbf{F1}} \\ \midrule
AntiMagn\_b      & 90.99                              & 93.15                     & 92.06                       & 85.92                              & 89.46                              & 87.65                                & 86.55                              & 81.78                              & 84.10                               \\
AntiMagn\_c      & 90.16                              & 94.39                     & 92.23                       & 82.11                              & 91.72                              & 86.65                                & 85.60                              & 83.55                              & 84.56                               \\ \midrule
AntiReal2\_b     & 77.13                              & 83.19                     & 80.05                       & 66.47                              & 86.39                              & 75.14                                & 83.69                              & 65.71                              & 73.62                               \\
AntiReal2\_c     & 83.83                              & 93.19                     & 88.26                       & 70.81                              & 92.10                              & 80.07                                & 79.57                              & 68.40                              & 73.62                               \\ \midrule
AntiVer\_b       & 96.05                              & 83.81                     & 89.51                       & 78.53                              & 46.53                              & 58.44                                & 89.57                              & 45.78                              & 60.59                               \\
AntiVer\_c       & 93.52                              & 88.88                     & 91.14                       & 78.81                              & 44.95                              & 57.25                                & 86.90                              & 46.12                              & 60.26                               \\ \midrule
CausFact0\_b     & 25.81                              & 08.26                     & 12.51                       & 62.79                              & 16.39                              & 25.99                                & 66.93                              & 19.47                              & 30.17                               \\
CausFact0\_c     & 18.33                              & 06.31                     & 09.39                       & 28.36                              & 7.79                              & 12.22                                & 67.20                              & 19.55                              & 30.29                               \\ \midrule
CausFunc0\_b     & 76.94                              & 30.66                     & 43.85                       & 66.27                              & 38.24                              & 48.49                                & 50.02                              & 32.86                              & 39.66                               \\
CausFunc0\_c     & 72.05                              & 34.67                     & 46.81                       & 71.04                              & 42.84                              & 53.44                                & 52.19                              & 32.27                              & 39.88                               \\ \midrule
CausFunc1\_b     & 91.15                              & 75.79                     & 82.76                       & 78.37                              & 70.94                              & 72.05                                & 89.00                              & 79.40                              & 83.93                               \\
CausFunc1\_c     & 89.40                              & 77.52                     & 83.04                       & 78.37                              & 71.84                              & 74.96                                & 87.63                              & 78.48                              & 82.80                               \\ \midrule
CausPredMinus\_b & 88.44                              & 68.09                     & 76.94                       & 82.31                              & 91.81                              & 86.80                                & 78.34                              & 62.86                              & 69.75                               \\
CausPredMinus\_c & 86.97                              & 69.70                     & 77.38                       & 82.57                              & 95.26                              & 88.46                                & 86.97                              & 71.05                              & 78.21                               \\ \midrule
Fact0\_b         & 80.10                              & 45.82                     & 58.30                       & 10.28                              & 6.65                              & 8.07                                & 19.40                              & 3.64
& 6.13                               \\
Fact0\_c         & 73.89                              & 49.14                     & 59.02                       & 10.59                              & 7.26                              & 8.61                                & 26.78                              & 4.63                              & 7.90                               \\ \midrule
FinFunc0\_b      & 0.00                              & 0.00                     & 0.00                       & 10.28                              & 6.65                              & 8.07                                & 0.00                              & 0.00                              & 0.00                               \\
FinFunc0\_c      & 0.00                              & 0.00                     & 0.00                       & 36.69                              & 12.36                              & 18.50                                & 0.00                              & 0.00                              & 0.00                               \\ \midrule
FinOper1\_b      & 98.44                              & 99.53                     & 98.98                       & 93.83                              & 99.16                              & 96.42                                & 92.20                              & 95.96                              & 94.04                               \\
FinOper1\_c      & 97.44                              & 99.69                     & 98.55                       & 64.52                              & 99.46                              & 96.93                                & 92.20                              & 95.96                              & 94.04                               \\ \midrule
IncepOper1\_b    & 78.54                              & 74.91                     & 76.68                       & 60.40                              & 62.15                              & 61.26                                & 96.30                              & 97.25                              & 96.77                               \\
IncepOper1\_c    & 82.10                              & 85.59                     & 83.81                       & 58.47                              & 66.09                              & 62.04                                & 71.41                              & 53.95                              & 61.46                               \\ \midrule
IncepPredPlus\_b & 95.53                              & 99.10                     & 97.28                       & 87.12                              & 90.50                              & 88.78                                & 71.41                              & 53.95                              & 61.46                               \\
IncepPredPlus\_c & 93.75                              & 98.85                     & 96.24                       & 88.21                              & 92.87 & 90.48                                & 95.42                              & 90.34                              & 92.81                               \\ \midrule
Magn\_b          & 40.35                              & 85.01                     & 54.72                       & 58.21                              & 82.08                              & 68.05                                & 49.24                              & 63.03                              & 55.27                               \\
Magn\_c          & 36.94                              & 97.22                     & 51.90                       & 64.44                              & 83.91                              & 70.94                                & 48.63                              & 63.92                              & 55.23                               \\ \midrule
Oper1\_b         & 38.11                              & 79.47                     & 51.90                       & 41.61                              & 59.48                              & 48.97                                & 34.81                              & 68.95                              & 46.26                               \\
Oper1\_c         & 37.11                              & 82.24                     & 51.14                       & 39.06                              & 72.75                              & 50.83                                & 32.85                              & 74.13                              & 45.52                               \\ \midrule
Real1\_b         & 41.22                              & 46.48                     & 43.69                       & 29.13                              & 25.30                              & 27.08                                & 37.55                              & 60.57                              & 46.36                               \\
Real1\_c         & 37.11                              & 82.24                     & 51.14                       & 29.16                              & 30.07                              & 29.61                                & 39.02                              & 63.45                              & 48.32                               \\ \midrule
Real2\_b         & 50.82                              & 42.43                     & 46.25                       & 59.61                              & 95.56                              & 73.42                                & 54.64                              & 54.53                              & 54.59                               \\
Real2\_c         & 50.66                              & 42.53                     & 46.24                       & 59.86                              & 94.65                              & 73.34                                & 55.67                              & 48.91                              & 52.07                               \\ \midrule
Ver\_b           & 80.97                              & 31.99                     & 45.86                       & 84.16                              & 85.30                              & 84.73                                & 89.17                              & 70.31                              & 78.62                               \\
Ver\_c           & 78.52                              & 32.74                     & 46.21                       & 84.16                              & 85.30                              & 84.73                                & 88.72                              & 70.17                              & 78.36                               \\ \bottomrule
\end{tabular}
}
\caption{Results breakdown per language and per LF, where, for each LF, we list individual results for base and collocate categorization.}
\label{tab:lf-results}
\end{table}

To obtain a more detailed picture, we report in Table \ref{tab:lf-results} the results of a run for the best performing models for each language and LF, for both of its collocation elements, the base (\_b) and the collocate (\_c). While there is certain consistency across LFs and languages, there are also notable cases of discrepancies. For instance, we see that Real2 (as, e.g., \textit{enjoy support}), Ver (as, e.g., \textit{legitimate demand}) and Magn (as, e.g., \textit{heavy smoker}) have been better captured in Spanish than in English and French. This can probably be explained by the number of unique instances of the LFs in our training / test data. For instance, in the case of Magn, the ratio between the total number of instances and the number of the unique number of instances  in the English test set is 16.8, while in the Spanish test set it is 31.8. In other words, our Spanish dataset contains less variety to express the meaning of intensification than English and French, and is thus easier to capture.
Conversely, the performance on Fact0 (as, e.g., \textit{an avalanche strike(s)}) is much better for English, which is likely due to the limitations of the training dataset: out of the 2,112 occurrences of Fact0 instances in total,  [\textit{el}] {\it avi\'on vuela} `the airplane flies' is counted 602 times.

Note the overall high figures of the recognition of the Magn and AntiMagn instances, and thus a clear distinction between these antonymic LFs, which is a well-known challenge \cite{rodriguez2016semantics,Wanner-etal17-ijl}.
In the case of AntiVer (as, e.g., \textit{illegitimate demand}), the figures are lower in the case of Spanish, which may again hint at the limitations of the Spanish dataset. For the prediction of the individual collocation items, in general, similar results are obtained for the base and collocate. However, some interesting outliers emerge. For instance, for the Spanish CausFact0 (as, e.g., \textit{start an engine}), the performance for the base elements (in our example, \textit{engine}) is more than twice as high as for the collocate elements (in our example, \textit{start}). We hypothesize that this is because most of the CausFact0 base elements in the Spanish dataset denote artefacts and the model learns to recognize them well. Finally, note that only the Spanish model is able to correctly identify a few FinFunc0 collocations (as, e.g., \textit{fire going out}), possibly due to the fact that Spanish contains less multiword expressions and certainly less phrasal verbs associated with this LF.

To understand whether there are obvious sources of confusion across LFs, and  whether we can attribute performance to frequency in the datasets, we plot in Figure \ref{fig:lf-analysis} confusion matrices, as well as the relationship between results and frequency. In English and French, Oper1 and Real1 are great sources of confusion for Real2, especially when it comes to categorizing Real2 collocates. However, this is not the case for Spanish. In this context, we need to keep in mind that Real1 and Real2 differ only with respect to their subcategorization pattern (in Real1, it is A0/A1, which is realized grammatical subject, and in Real2, it is A2) and that the semantic difference betweeen Oper and Real is rather fine. Still, for Spanish this difference is captured, while for English and French it is not. This is similar for the 
distinction between  CausFact$_i$ / Oper$_i$ and Real$_i$. 
Why the confusions are minor for Spanish requires a deeper analysis. We can also see that Magn and Oper bases are often confused in French, but not in English and Spanish. This might be due to parsing and PoS tagging errors. Finally, in the lower  part of Figure \ref{fig:lf-analysis}, we see 
that for English, there is a clear correlation between results and LF frequencies ($\rho$=0.76),  followed by French ($\rho$=0.46) and, finally, Spanish ($\rho$=0.38), where we also find highest dispersion across all F1 bins.




\section{Conclusions and Future Work}
\label{sec:conclusions}

We have proposed an architecture for joint collocation extraction and lexical function typification by explicitly encoding syntactic dependencies in the attention mechanism. 
Our experiments show that our proposed architecture drastically improves over its language model-only counterparts, and that joint multilingual training is a promising direction for less resourced languages. 
For the future, we would like to extend these experiments to other languages and explore zero or few-shot prompt-based methods.

\section*{Acknowledgements}
Many thanks to Beatriz Fisas, Alba T\'aboas, and Inmaculada L\'opez for their help with the datasets.
We would also like to thank the anonymous reviewers for their very helpful comments. The work by Alexander Shvets and Leo Wanner has been supported by the European Commission in the context of the Horizon 2020 Research Program under the grant numbers 825079 and 870930. Alireza Mohammadshahi is supported by the Swiss National Science Foundation~(grant number CRSII5-180320).

\bibliography{custom}
\bibliographystyle{acl_natbib}

\end{document}